# Creating a morphological and syntactic tagged corpus for the Uzbek language

Maksud Sharipov [1], Jamolbek Mattiev [1], Jasur Sobirov [1], Rustam Baltayev [2]

[1] Urgench State University, Khamid Alimdjan 14, Urgench, 220100, Uzbekistan
[2] Urgench Branch of Tashkent university of Information Technologies Named After Muhammad al-Khwarizmi, 110, Al-Khwarizmi str, Urgench, 220100, Uzbekistan

**Abstract**
Nowadays, creation of the tagged corpora is becoming one of the most important tasks of Natural Language Processing (NLP). There are not enough tagged corpora to build machine learning models for the low-resource Uzbek language. In this paper, we tried to fill that gap by developing a novel Part Of Speech (POS) and syntactic tagset for creating the syntactic and morphologically tagged corpus of the Uzbek language. This work also includes detailed description and presentation of a web-based application to work on a tagging as well. Based on the developed annotation tool and the software, we share our experience results of the first stage of the tagged corpus creaton.

## 1. Introduction

Nowadays, the Natural Language Processing (NLP) field is developing rapidly and is playing an important role to solve the problems in the scientific, economic, and cultural fields. NLP also covers industries such as business data analysis, web application development, corpus linguistics, computer science, as well as artificial intelligence. The majority of the information available on the Internet is textual, therefore, obtaining the necessary information through the analysis of textual data, through various techniques, such as morphological and syntactic analysis of such texts, are becoming main fields of interest in NLP.

To date, there are many language corpora of most spoken languages, some of the very early works and also popular ones are the Brown corpus [1], and the International Corpus of English and the British National Corpus [2]. At present, practical research is underway in the field of corpus linguistics to create language corpus for various purposes. The usefulness of corpora for linguistic research works is provided by the creation of tagged sub-corpus in these corpora [3].

Some research works have been done to create tagged corpora for the Uzbek language, for example: [4,5] which provides information on the basic requirements and principles of linguistic annotation for text processing in the creation of the electronic corpus of the Uzbek language, and the results of theoretical and practical research on morphological tagging and morphological analyzer construction using FST technology.

Due to the lack of language resources in Uzbek language, there are difficulties in solving NLP problems. To solve NLP problems, we need a morphologically and syntactically tagged corpus. To date, the lack of an open source morphologically and syntactically tagged corpus for the Uzbek language makes it difficult to conduct research in the field of linguistics. There are 12 word classes in the Uzbek language. A word can be polyfunctional depending on the state of its realization in the sentence and the semantic valence of the N-gramm words [6]. The typical approach for most NLP applications using tagged corpora consists of the creation of a corpus through manual annotation and then training a machine learning model [7]. To solve the above-mentioned issues, we aimed to create an open source tagged corpus in this research. The goal is to build a supervised tagger using the tagged corpus which is being created. Typically, a pre-tagged corpus using a tool is required to create supervised taggers [8].

The importance of the proposed work lies beneath the complex structure of the tagset built, and the tool to annotate given texts to create a tagged corpus, which in turn will be used for the upcoming work of tagger tool for Uzbek, to train sequence labeling language model.

## 2. Related work

Since a morphological and syntctic tagset and tagged corpus is one of the fundamental must-have resources and one of the first steps of creating resources for languages, all the well-resourced languages can be said to have their tagsets and tagged corpora developed at some point. All the languages in use differ from each other with their syntax, morphology and phonetics, but at teh same time, majority of them have a similar constructive structure, which allows linguists to create multilingual resources and tools. In an attempt to a creation of a multilingual tagset that can be used by as many languages as possible, there has been a work by Google research to create a universal POS tagset [9], which presents a tagset that was obtained by mapping similar features of 22 languages together. This universal POS tagset is now used by many languages as the base of their tagset, which is then extended by more tags that encode language-specific features. This universal POS tagset is also used by the Universal Dependencies (UD) project [10], one of the fastest growing multilingual tagged NLP data platform that has data over 130 languages.

On the topic of a similar work done on Uzbek language, the first work that presented the morphological tags list and the morphological tagger [11] presented a tool created in Prolog. But the problem with the work was that it only covered main parts of speech in Uzbek text, and was missing many tags to deal with complex words.

In [12], the issue of tagging the Uzbek language corpus was considered. Authors proposed 14 POS tags, that is, almost one tag is created for each word class, but in Uzbek language each word class is divided into several types in terms of meaning and structure. In our approach, we took into consideration those issues and created the expanded tagset by deeper analysis. The novel tagset allows us to analyze the text in depth from a semantic point of view. In [13], the importance of rule-based and stochastic tagging methods for the Uzbek language is discussed. The need of a tagged corpus for the Uzbek language is indicated and the occurrence of words in sentences with different functions is described, however, authors did not provide any morphological or syntactic tagset which can be used for tagging.

There are very limited amount of NLP work done on Uzbek, some of the important ones include Sentiment analysis datasets [14,15], cross-lingual word embeddings over closely-related Turkic languages [16], stopwords dataset [17], Stemmer for Uzbek verbs [18], as well as recent neural transformer based (BERT) language model [19] which was trained on a big raaw Uzbek text. Although there is a big amount of scientific works published claiming that they have contributed to the Uzbek NLP, the quality of works, be it a language resource, or a tool, is nowhere near that amount. This statement about some scientific works which claim they have done something, but not providing an open-source code or the data itself, are mentioned as "zigglebottom" papers in a recent work done on Uzbek [20].

Regarding related works done on similar languages, there has been a work done on the Kazakh language [21], which syntactic and POS tags have been developed to create a tagged corpus. The authors produced 36 morphological tags and 9 syntactic tags and developed an annotated corpus which consist of 613 511 words based on their tagset.

## 3. Proposed methods

We know that corpus can be used as a basis in many fields and scientific processes. Considering that corpus texts come in different genres and categories, it is easier to use a corpus if each word is accompanied by its morphological and syntactic classification (which group of words it belongs to and which syntactic function it belongs to in a sentence) provided. This is the process of text interpretation.

This work explains the question of how it is done for the Uzbek language specifically.

### 3.1. Tag list development

First of all, for the interpretation of Uzbek words, tags (explanations) are needed, consisting of abbreviations expressing morphological and syntactic meanings. Table 1 shows some of the morphological tags which we used in below examples (Detailed information about whole morphological and syntactic tag lists can be found at [22])

**Table 1**
Description of selected morphologic tags, with description and example words for each part of speech.

| Name | Tag | Description | Example |
|---|---|---|---|
| NOUN | SOT | Personal noun | Teacher (o'qituvchi) |
|  | NOT | Object noun | Bag (sumka) |
|  | JOT | Place noun | Village (qishloq) |
|  | MOT | Abstract noun | Love (muhabbat) |
| ADJECTIVE | XSF | Peculiarity-state adjective | Hard (qattiq) |
|  | RSF | Color adjective | White, black (oq, qora) |
| PRONOUN | KOL | Personal Pronouns | I (men) |
|  | KROL | Demonstrative Pronouns | This (shu) |
| ADVERB | HRV | Adverbs of manner | Rapidly (tez) |
|  | MIRV | Adverbs of quantifiers | A lot (ko'p) |
|  | PRV | Adverbs moder fier of time | Before (avval) |
| VERB | SIFL | Past participle verb form | Gone, seen (borgan, ko'rgan) |
|  | HFL | Infinitive form of the verb | Going, saying (borish, aytish) |
|  | SFL | Original verb form | See, read (ko'rdi, o'qidi) |
|  | KFSQ | Auxiliary verb combination | Fell in love (sevib qoldi) |
|  | 1B | First person singular | I am flying (uchyapman) |
|  | 3B | Third person singular | He said (aytdi) |
|  | 2K | Second person plural | You worked (ishladingiz) |
|  | OTZ | Past simple tense | I said (aytdim) |
|  | KEZ | Future simple | I will tell (aytaman) |

It is known from linguistics that the morphology field studies words, their categories and morphological features. The morphological analysis indicates to which category the word belongs, its nominal form and suffixes. The above table shows the word-class, its conditional abbreviation (POS tags), the description of the category, and examples. For example, if we take the verb word-class, here are provided 9 POS tags belonging to this category and some examples of them. The examples above are just a few examples of common morphological tags that belong to the noun, adjective, pronoun, adverb and verb family. Information about the whole tagset is shown in Table2.

It can be seen from the Table 2 that we created 102 morphological tags in total for the Uzbek language. For example: 22 POS tags were developed for the noun word-class, 10 POS tags for the adjective word-class, 11 POS tags for pronouns and so on. Similarly, tags are used in the parsing of words.

**Table 2**

Detailed information about the whole POS tagset. Number of POS tags created for each word class are presented.

| Word class | Part of speech tags |
|---|---|
| NOUN | 22 |
| ADJECTIVE | 10 |
| NUMBER | 11 |
| PRONOUN | 11 |
| ADVERB | 10 |
| VERB | 18 |
| CONJUNCTION | 8 |
| HELPERS | 1 |
| PARTICLE | 6 |
| INTERJECTION | 2 |
| IMITATIVE WORD | 2 |
| MODAL WORD | 1 |
| **Total:** | **102** |

We developed the comprehensive morphological tagset for the Uzbek language by deeper analysis for in-depth morphological tagging of Uzbek words. Similarly, the syntactic tagset was also created in this research for syntactic tagging of Uzbek words. The Table 3 lists 14 syntax tags and examples of their usage:

**Table 3**

Detailed information about the whole syntactic tagset. For each syntactic tag, there is a small description and an example is given.

| Name | Syntactic tag | Description | Example |
|---|---|---|---|
| SUBJECT | EG | Subject | Brother came home |
| PREDICATE | OK | Noun Predicate | Urgench is a beautiful city |
|  | FK | Verb Predicate | Salim came |
| ATTRIBUTE | QA | Genetive Attribute | My brother's face |
|  | SA | Adjectival Attribute | Good students were rewarded |
| OBJECT | VL | Indirect Object | We talked about home |
|  | VS | Direct Object | The knife cut my hand |
| MODIFIERS | VH | Condition Modifiers | He agreed out of desperation |
|  | PH | Time Modifier | He finished work in the evening |
|  | OH | Place Modifier | He wants to live in Tashkent |
|  | SH | The Reason Modifier | Kasallangani uchun kelmadi |
|  | MH | The Aim Modifier | He deliberately does not enter the building |
| EXCLAMATION | UN | A person or object that is focused on speech | Anwar, look at me |
| THE ENTRY WORD | KR | The entry word | Unfortunately, he returned home |

The Syntax section studies phrases and sentences. Syntactic analysis of a sentence analyzes the parts that make it up: the relationship of 5 parts of speech (there are 5 parts of speech in Uzbek language, namely: subject, predicate, attribute, object and modifiers.There are also some parts of speech that do not interact with the those parts of speech, which are called "EXCLAMATION" and "THE ENTRY WORD") and the parts of speech that do not interact with the parts of speech. Syntactic analysis is not only a scientific aspect of linguistics, but also plays an important role in the attractiveness of discourse and the fluent formation of the text. Table 3 provides the names of the parts of speech, their conditional abbreviations, explanations of the parts of speech, and related examples. For example: if we look at a "predicate" word-class, we can see that it has 2 different types (OK and FK), their names (Noun Predicate and Verb Predicate), and examples of their usage.

### 3.2. Developed algorithm for tagging

Part of speech is much more complicated than simply comparing words to word classes. Because POS and syntactic tagging are not easy. A single word can serve as a different word class in different sentences based on different contexts [3]. So far, there is not enough tagged corpus for the Uzbek language to create machine learning algorithms, so the main goal of our research is to develop an algorithm for tagging texts and to develop a web-based tagging system. All the tags and the tagger proposed in this use the official Latin alphabet as a default script, but the problem with texts in Uzbek language is that the old Cyrillic script is equally popular all in official written documents, literature, as well as internet websites. The texts that appear in Cyrillic are pre-processed using available tools, such as web-based transliterator Savodxon[2], or a machine transliteration Application Programming Interface (API) [23], before being fed as an input to the tagger. The steps for syntactic and morphological tagging of texts is shown in Figure 1.

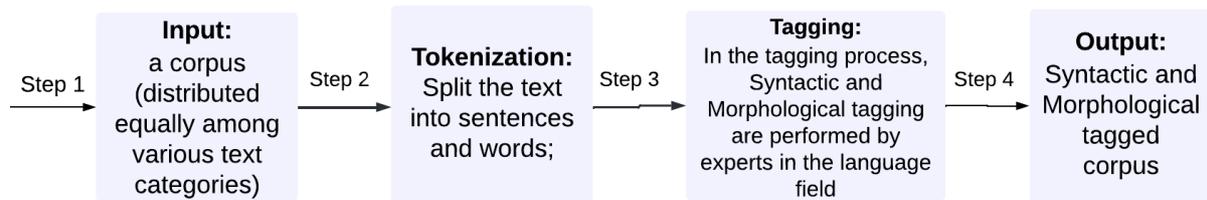

**Figure 1:** Syntactic and Morphological tagging of texts

According to Figure 1, we developed a web-based tagging application which can be found at [24]. To utilize the application, user has to follow the following steps:
- Registration of experts who perform the syntactic and morphological tagging of texts;
- Extracting the texts from corpus and splitting them into sentences as well as words;
- Sending the selected sentence to the user interface;
- Writing the result to the file with the user ID number of the tagged sentence;
- Producing the final result after each sentence is tagged;
- Writing the result to the corpus in XML or TXT format;

### 3.3. Developed tagged corpus

The following syntactic and morphologically tagged corpus is created based on the developed web application. Let's see how tags are used in several sentences (a - morphological tags; b - syntactic tags):

1. **I opened the window to get some fresh air (Men biroz toza havo olish uchun derazani ochdim)**
  a) Men/KOL biroz/MIRV toza/XSF havo/MOT olish/HFL uchun/KM derazani/NOT ochdim/SFL/1B/OTZ
  b) Men/EG biroz/PH toza+havo+olish+uchun/MH derazani/VS ochdim/FK
2. **Anvar suddenly came to the door**
  a) Anvar/SOT to'satdan/HRV eshik/NOT yoniga/JOT keldi/SFL/3B/OTZ
  b) Anvar/EG to'satdan/VH eshik+yoniga/OH keldi/FK

---

[2] Savodxon machine transliterator: https://savodxon.uz/

3. **Only today you can buy a car at this price, tomorrow new price will be set (Faqat bugun ushbu narxda mashina xarid qila olasiz, ertaga yangi narx qo'yiladi)**
a) Bugun/PRV ushbu/KOL narxda/MOT mashina/NOT xarid+qila+olasiz/SFL/2K/KEZ, ertaga/PRV yangi/XSF narx/MOT qo'yiladi/SFL/3B/KEZ
b) Bugun/PH ushbu/SA narxda/VL mashina/VS xarid+qila+olasiz/FK, ertaga/PH yangi/SA narx/EG qo'yiladi/FK
4. **A snake cannot move on a flat surface (Ilon yassi yuzada harakatlana olmaydi)**
a) Ilon/NOT yassi/XSF yuzada/JOT harakatlana+olmaydi/KFSQ
b) Ilon/EG yassi/SA yuzada/OH harakatlana+olmaydi/FK
5. **You have to take into consideration the performance characteristics rather than its price when you are buying a mobile phone (Siz mobil telefon sotib olayotganingizda uning narxiga emas, ishlash xususiyatlariga e'tibor qaratishingiz kerak)**
a) Siz/KOL mobil/XSF telefon/NOT sotib+olayotganingizda/SIFL uning/KROL narxiga/MOT emas, ishlash/HFL xususiyatlariga/MOT e'tibor+qaratishingiz+kerak/HFL/2K
b) Siz/EG mobil/SA telefon/VS sotib+olayotganingizda/PH uning/QA narxiga/VL emas, ishlash/SA xususiyatlariga/VL e'tibor+qaratishingiz+kerak/OK

Let's now briefly explain how these symbols are used. Each interpreted word is followed by a forward slash (/) followed by a shorthand tag(s) indicating the morphological or syntactic nature of the word (for example, / is NOT followed by an object). The number of tags that can be placed after a word can be more than one (came/ SFL/3B/OTZ). Figure 2 shows that the tagging process of words/sentences in newly developed web-based application.

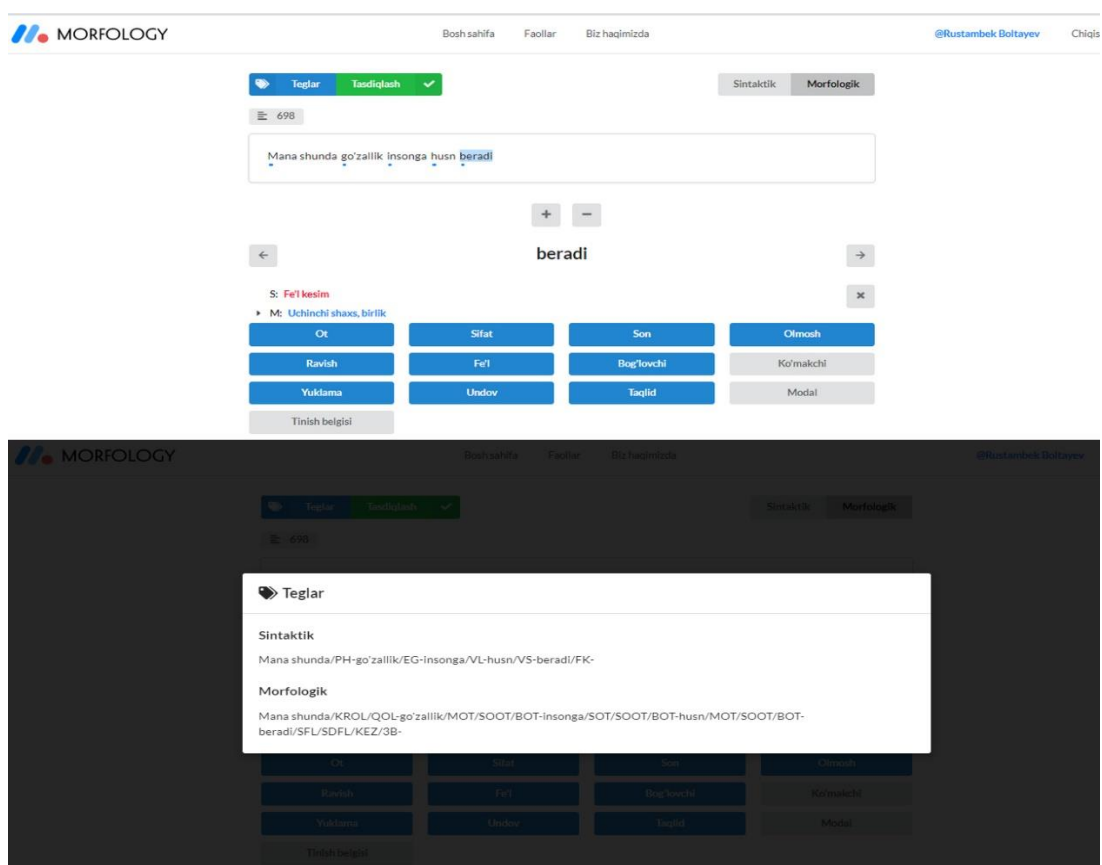

**Figure 2**: Example of tagging process in web-based application

Application is easy to follow: (1) the text is inserted into the database; (2) you have to select the tagging type: syntactic or morphologic (grey buttons in right-top corner); (3) the tagging process is performed word-by-word, so you should choose the tag name from the developed tagset shown below the text. A tagged word is added to the tagged text by clicking +/- buttons. (4) Once you tag all the words in the text, you can add the tagged text into the tagged corpus by clicking the "Tasdiqlash" ("confirm", green color) button.

## 4. Results and evaluation

A morphologically and syntactically tagged corpus for the Uzbek language was created by using the developed tagset and program. More than 1,200 sentences and more than 10,000 words from texts in different fields (a total of more than 15 categories), such as literature, technology, sports, psychology, politics, society, medicine, religion and philosophy, have been tagged and the tagging process is still going on. Our main goal is to develop the largest tagged corpus for the Uzbek language. The current tagged corpus created by us can be used by researchers as a dataset to solve NLP tasks in Uzbek language. Since there is no open-source tagged corpus for the Uzbek language, this research work can be considered a novel and important contribution in the NLP field for the Uzbek language.

## 5. Conclusion and future work

In this paper, a part of speech tagset which is required to create a tagged corpus was developed for the Uzbek language. A web-based annotation tool for tagging texts based on the developed tagset were created. Using the created program, the texts are tagged by experts. Using the syntactically and morphologically tagged corpus of the Uzbek language created by us, it is possible to solve such problems as Named entity recognition, statistical language modeling, text generation pattern identification, machine translation and syntactic analysis. The tagged dataset is now being expanded. In the future work, it is planned to build automatic tagging algorithms (named: Uzbek tagger) for the Uzbek language with machine learning using a tagged dataset.

The created annotation tool can be used for other Turkic languages as well, for which it is necessary to place the tag set of this language in the application. The process of applying the algorithm to other Turkic languages can be carried out by using most of the available tags, plus some language-specific tags regarding the target language.

## 6. Acknowledgements

The author (Jamolbek Mattiev) gratefully acknowledges the European Commission for funding the InnoRenewCoE project (Grant Agreement #739574) under the Horizon2020 Widespread-Teaming program and the Republic of Slovenia (Investment funding of the Republic of Slovenia and the European Union of the European Regional Development Fund). Jamolbek Mattiev is also funded for his Ph.D. by the "El-Yurt-Umidi" foundation under the Cabinet of Ministers of the Republic of Uzbekistan.